# A Formally Verified Fail-Operational Safety Concept for Automated Driving

Yuting Fu
Systems Innovations
NXP Semiconductors
Eindhoven, The Netherlands
yuting.fu@nxp.com

Andrei Terechko
Systems Innovations
NXP Semiconductors
Eindhoven, The Netherlands
andrei.terechko@nxp.com

Jan Friso Groote
Department of Mathematics and Computer Science
Eindhoven University of Technology
Eindhoven, The Netherlands
j.f.groote@tue.nl

Arash Khabbaz Saberi
Department of Mathematics and Computer Science
Eindhoven University of Technology
Eindhoven, The Netherlands
a.khabbaz.saberi@tue.nl

## ABSTRACT

Modern Automated Driving (AD) systems rely on safety measures to handle faults and to bring vehicle to a safe state. To eradicate lethal road accidents, car manufacturers are constantly introducing new perception as well as control systems. Contemporary automotive design and safety engineering best practices are suitable for analyzing system components in isolation, whereas today's highly complex and interdependent AD systems require novel approach to ensure resilience to multi-point failures. We present a holistic safety concept unifying advanced safety measures for handling multiple-point faults. Our proposed approach enables designers to focus on more pressing issues such as handling fault-free hazardous behavior associated with system performance limitations. To verify our approach, we developed an executable model of the safety concept in the formal specification language mCRL2. The model behavior is governed by a four-mode degradation policy controlling distributed processors, redundant communication networks, and virtual machines. To keep the vehicle as safe as possible our degradation policy can reduce driving comfort or AD system's availability using additional low-cost driving channels. We formalized five safety requirements in the modal μ-calculus and proved them against our mCRL2 model, which is intractable to accomplish exhaustively using traditional road tests or simulation techniques. In conclusion, our formally proven safety concept defines a holistic design pattern for designing AD systems.

## 1 Introduction

Safety of autonomous vehicle passengers and other road users depends on Automated Driving (AD) systems. To increase traffic safety, governments and the automotive industry formulated Vision Zero [3] with the goal of completely eradicating road fatalities in future. To design an AD system, automotive safety engineering involves a reliable product lifecycle management, safety analysis techniques, and design patterns. State-of-the-art safety standards and guidelines (ISO 26262 [1], ISO/PAS 21448 SOTIF [2], SaFAD [12], RSS [14], IEEE P2846 [15], etc.) are developed to reduce risks due to system malfunction, fault-free hazardous behavior and wrong driving decisions. Our work contributes by providing a holistic fail-operational design pattern and exploring the promising practice of formal verification applied to a novel highly resilient degradation policy.

The ISO 26262 standard [1] focuses on system malfunctioning due to random hardware faults and systematic software faults. According to this standard, the required integrity of AD elements is indicated by the Automotive Safety Integrity Levels (ASIL). Elements with higher ASILs are required to be more reliable, but also more costly to develop. Commercially viable AD systems favor cost-effective solutions with fewer high ASIL elements. ASIL decomposition [1] process can help reducing required ASIL, given that there is sufficient architectural, hardware, or software independency [4].

Traditional automotive systems required only fail-safe mechanisms, however, AD application (especially automation level 3+) require fail-operational capabilities. Developing fail-operation AD systems requires complex safety analysis such as dependent failure analysis and ensuring resilient to multiple-point-faults. In this paper, we analyze handling dual-point and some triple-point faults. Our proposed solution provides fail-operational capabilities while limiting the number of high ASIL elements to optimize the cost [12].

Furthermore, the AD applications need situational awareness in the absence of faults to monitor its own performance limitations as specified in the SOTIF [2]. A common safety measure against performance limitations is to follow a degradation policy depending on the Operational Design Domain (ODD) [12][27]. An ODD specifies conditions under which the AD application is designed to correctly control the vehicle such as the road type, speed range, weather and light conditions [28]. Note that AD components may have different performance in different ODDs (e.g. components that are based on machine learning), and may not always be safe to use.

Formal verification techniques [23] are not strictly required by ISO 26262, which rather focuses on traditional methods, such as failure analysis techniques, testing and validation. Unfortunately, the traditional techniques are not exhaustive and not time-effective [13], resulting in late identification of safety issues in the development or in the field. In contrast, formal model checking reduces costly specification errors by exhaustively proving safety properties [16]. Noteworthy that an IEEE P2846 industrial working group [15] is currently defining a formal model for automated decision making using mathematical formulas of vehicle kinematics.

We propose a safety concept based on a Distributed Safety Mechanism (DSM) for automated driving with various degraded modes capable of handling multiple-point faults. To verify our approach, we model the DSM behavior using the formal specification language mCRL2 [24]. The safety requirements of the DSM model are defined, formalized, and formally verified.

This paper makes the following contributions:
1) A DSM based, fail-operational safety concept (including behavioral specification) to handle multiple-point faults and performance limitations using degraded modes and allocation of fault models and monitoring techniques to different safety layers of the DSM.
2) A formal specification of the DSM model using the mCRL2 language and formally verified safety requirements for the DSM degraded modes.

The rest of the paper is organized as follows. In Section 2 we give an overview of related work. Then we describe the DSM architecture in Section 3. We discuss allocation of fault models and monitoring techniques to our DSM safety layers in Section 4. We elaborate on the fault handling and degraded modes behavior in Section 5. The DSM model and the formal verification process are presented in Section 6. Finally, we give conclusion in Section 7.

## 2 RELATED WORK

In this section we present an overview of related AD systems' safety architectures and the application of formal modeling in the automotive domain that are most relevant to our work.

### 2.1 Safety Architectures for Automated Driving

A distributed safety mechanism using hypervisors and different middleware software stacks was proposed in [4]. The safety mechanism was implemented on a hardware-in-the-loop setup with an automated driving simulator. The proposed DSM in [4] is evaluated by proof-of-concept experiments on the hardware-in-the-loop simulation to be able to detect single-point faults in the AD system and safely stop the vehicle when necessary. Our work inherits the layered architecture from [4] and the employment of software middleware [19] and hypervisors. However, in contrast to [4] which focused on a proof-of-concept implementation of the DSM architecture, our work refines the safety concept to enable more degraded modes. Furthermore, we model and formally verify the logic of the DSM fault handling behavior.

A scalable architecture for AD system is proposed in [6]. In this architecture there is one primary ASIL channel that provides advanced autonomous driving functionalities and comfort for the passengers. On the same platform there is a second ASIL channel with less advanced functionalities. But it has, for example, safer trajectory plans and is therefore safer than the primary channel. A selector is implemented to switch between the two ASIL channels. Besides, a fail-degraded channel is implemented on a separate platform with access to redundant braking and steering. Scalability and integration are the main focus of this design. Compared to [6] that relies on redundant actuators for degraded operation, our DSM relies on redundant communication networks, middleware stacks, hypervisors, and different monitoring techniques to enable various degraded forms of AD system operation.

A fail-operational architecture, which remains operational in the presence of faults [9], is proposed in [7]. It has two dual-channel domain controllers. Each of the four channels consists of a chain of sensor systems, AI-based data processing and control, and actuator systems. Note that only the data processing and control system is independent for each channel, all the other subsystems are shared among the channels. The secondary channel in our DSM is more isolated in this sense because it has a different AD system functionality implementation and its data comes from dedicated safety sensors. In [7], each channel has multiple monitors and they all report to a global safety/fault manager in the same channel. This centralized channel safety manager is responsible for fault containment and response functions in the channel. Similar systems are the centralized health monitor implemented in the Baidu Apollo AD framework [22], and a centralized AD system Mode Manager described in [12]. On the contrary, our DSM supports distributed monitors in all safety layers. The fault detection is reported not to a centralized monitor, but to a diagnostic message topic provided by the middleware software. Another interesting point [7] made is that time required to find the minimal risk safe stop is indeterminant and could take several minutes. Our DSM copes with this challenge by providing further degradation even during certain degraded modes of the AD system.

Several system-level safety systems for truck platooning architectures applying the safety executive pattern [10] were discussed in [8] and [9]. The described architectures implement two heterogeneous channels running on different platforms. Different voting mechanisms are used to arbitrate between the two channels. This heterogeneous duplex pattern [10] is fail-operational in case of a single failure in one of the channels [8]. In the DSM we also implement a heterogeneous safety channel next to the nominal channel and extend fault handling to multiple-point faults.

In [11] a monitor/actuator architecture for autonomous vehicles, also known as the doer/checker architecture, is described. This architecture depends on an ideal monitor to detect a failure in the primary system. Then a simple and high-integrity failover system can bring the vehicle to a safe state.

The multiple degraded modes of operation enabled by our DSM are similar to the failover mission described in [11]. The advantage of the DSM is that by making use of the layered architecture and different monitoring techniques, we can still guarantee fail-operation even if the monitor becomes faulty.

A functional generic architecture for AD systems is proposed in [12]. There are one or multiple AD system Mode Manager modules that switch the AD system from nominal to degraded operation based on information received from multiple monitors. While [12] describes its framework at a fairly high-level, we present our DSM with both sufficient low-level hardware details and high-level safety mechanism behavior and allocation of fault models and mitigation techniques to the DSM architecture.

## 2.2 Modeling and Formal Methods for Automated Driving

The Responsibility-Sensitive Safety (RSS) [14] and the related IEEE P2846 working group [15] define a mathematical model for safe driving decisions. The model enforces longitudinal and lateral distance between the vehicles, the right of way, and evasive maneuvers. It defines a "Safe State" designed to prevent the autonomous vehicle from being the cause of a road accident from the legal perspective. Note, that RSS assumes that the AD system never malfunctions and has no performance limitations, for example, the perception subsystem perfectly detects all road users. Furthermore, the RSS model is technology neutral, omitting safety-critical effects in the System-On-Chip, networks, and software. Our safety concept incorporates RSS as a crucial component for driving decision making. To resolve RSS limitations our concept uses other techniques; for example, system monitoring and reaction mechanisms to address the AD system malfunctioning.

In [17] a simplified fail-operational model of a brake by wire system is given using SysML [18] diagrams and the transitions are analyzed using the activity diagrams. There is arbitration logic to activate the fallback channel in case of a failure in the nominal channel. Faults in one channel are assumed to be independent from faults in the other channel. There is one emergency operation mode and a driver takeover mode in case of a failure. Both channels in [17] are fail-silent, which means they shall become silent and produce no output at all in case of a failure without interfering any components, such that failure propagation is prevented at the component level. Our safety concept assumes only one safety layer to be fail-silent and relies on Virtual Machines (VM) to prevent fault propagation. Also, more degraded modes are implemented in the DSM to handle multiple-point faults.

Fault Tree Analysis (FTA) is used in [17] to deduce a fault tree failure model based on the state machine and the activity diagrams of the system. The result is a safety framework compliant with ISO 26262 [1]. Although no formal methods are applied in [17], the necessity of proving the correctness of the arbitration logic and deadlock-free property of the system is emphasized and formal verification is proposed as the solution to prevent failure in the logic of the state machine controlling fail-operational driving.

A framework to capture the relation between AD system design and functional safety is proposed in [13]. The work discusses the strengths and benefits of formal verification in the automotive domain and suggests applying formal methods to help address concerns about unintended functions and emergent behavior, which are not covered by ISO 26262 or SOTIF.

## 3 ARCHITECTURE OF THE SAFETY MECHANISM

In the following subsections we detail the DSM architecture and component functionalities in our safety concept.

### 3.1 Functionalities of the Distributed Safety Mechanism

Figure 1 illustrates the high-level architecture of our DSM. Note that we do not limit the architecture to any specific implementation, but only give implementation examples for illustrative purposes. The functionalities of each component are described below.

a) AD and ODD Sensors: There are AD sensors and ODD sensors in the AD system. The AD sensors provide data via the primary network to particular AD function modules on the Function (FUN) layer. For example, cameras and radar sensors provide data to the perception function module, while the global navigation satellite system feeds the localization module. At the same time the AD sensor data are monitored by the Sensor and Function Monitor (SFM) layer. The intention of the ODD sensors is to provide input via the primary network to an ODD checker integrated in the SFM layer.

b) Safety Sensors: There are hot standby sensors producing data for safe maneuver via the secondary network, which are received and monitored by the Vehicle Safety Mechanism (VSM) layer.

c) Function (FUN) Layer: In the FUN layer the AD function modules, such as localization, perception, path planning, etc. are running. The FUN layer has access only to the primary network channel. One or multiple modules with related functionalities run in the same VM. Note, that safety-related decision making algorithms, such as RSS [14], are typically integrated in the path planning module, as it is the case in Baidu Apollo AD platform [22].

d) Middleware: Middleware is software that enables the various components of a distributed system to communicate reliably and manage data [11]. Communication reliability and visibility are guaranteed by middleware Quality of Service (QoS) policies, such as real-time deadline, priority, and reliability [21]. Note that the middleware communication protocol has to be free of single point of failures, see, for example, the Data Distribution Service (DDS) standard [21]. In our DSM, we assume information is shared on top of the reliable middleware that uses the publish-subscribe communication pattern [19].

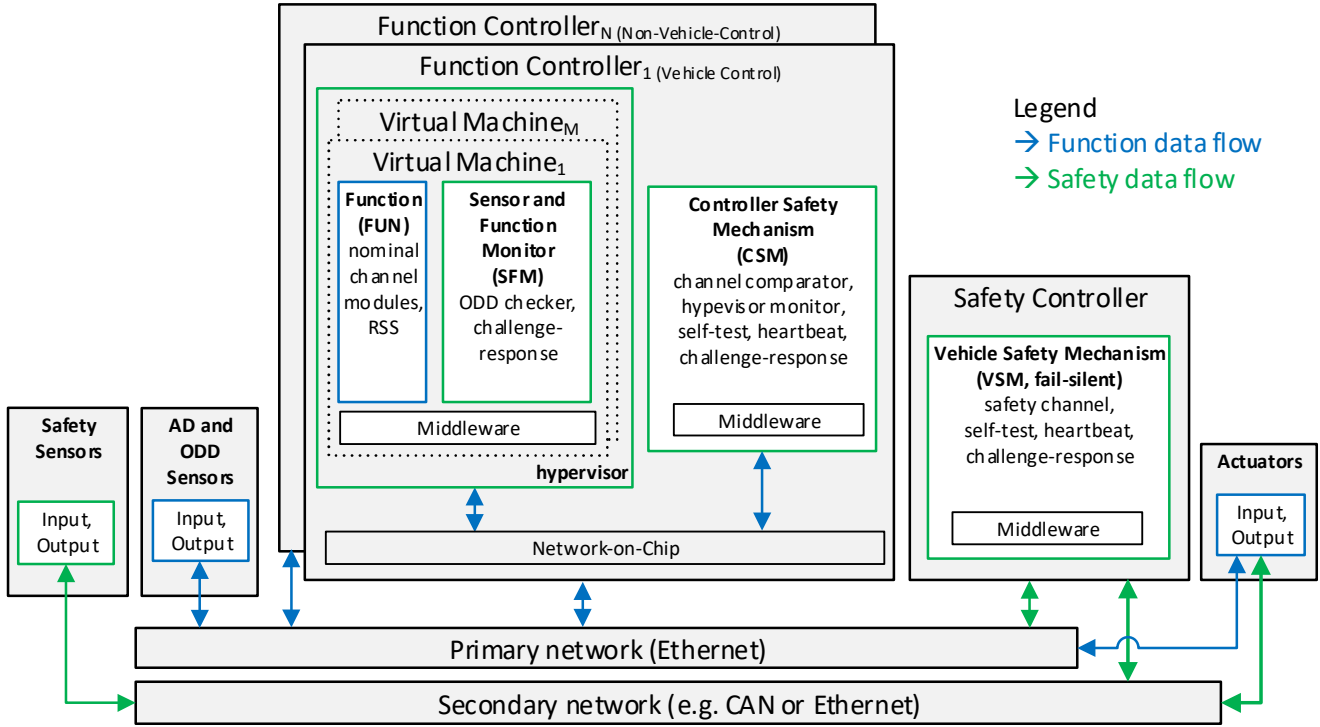

**Figure 1: The architecture of an Automated Driving System with the Distributed Safety Mechanism**

e) Sensor and Function Monitor (SFM) Layer: Each VM has an SFM layer. The SFM monitors the status of the FUN layer in the same VM. Furthermore, it detects faults in the corresponding AD sensor inputs and outputs, such as timing jitter and out-of-range message data. The SFM layer also acts as an ODD checker. Based on the ODD sensor outputs, it detects when the driving situation changed [28] and triggers necessary AD system mode transitions to avoid unsafe usage of components with limited performance, such as neural networks in the AD system perception subsystem.

f) Controller Safety Mechanism (CSM) Layer: Each System-On-Chip (SoC) in Figure 1 is annotated as Function Controller and has a dedicated CSM layer. The SoC with vehicle control functionalities is termed as the Vehicle Control (VC) function controller. The others are called Non-Vehicle-Control (NVC) function controllers. NVC CSMs only monitor their local SFM layers, hardware, and hypervisors; the VC CSM monitors all the VC and NVC function controllers and the VSM layer. The CSM layer has access to the primary network channel and can send control commands to the vehicle actuators. Note that the CSM layer can compare channels' outputs to identify disagreement between the nominal and safety channels.

g) Vehicle Safety Mechanism (VSM) Layer: The VSM layer runs on a separate safety controller. It monitors the VC CSM, the safety sensor data, and the two off-chip networks. The VSM layer can maneuver the vehicle via the secondary network using the safety sensor data. In addition, we assume the VSM layer has access to the power management ICs.

h) Network-On-Chip (NoC). The NoC is a low latency communication interconnect for hardware Intellectual Property (IP) modules on an SoC.

i) Primary Network: This is the high performance network channel for nominal control of the AD system.

j) Secondary Network: This hot standby of the primary network is a heterogeneous high-performance redundant communication channel for safety control of the vehicle.

### 3.2 Cost-Effective Fail-Operational Architecture

In the scope of this paper, we are not able to provide empirical evidence regarding cost-effectiveness of our proposed approach. However, we are able to qualitatively compare our approach to other well-known patterns. The Triple Modular Redundancy (TMR) design pattern [10] is commonly used in aerospace systems, such as Boeing 777 airplane flight control system [30]. TMR consists of three costly redundant channels and a reliable voter. This pattern provides high degrees of fault resilient and can be fail-operational, however, TMR is considered too expensive for AD systems. Our proposed pattern adds simpler and cheaper emergency and safety channels to the nominal channel. Based on the system health status and environmental awareness, the DSM employs a degradation policy to switch between the channels, which is described in Section V. The simplest emergency channel does not use sensors at all and merely stops the vehicle. Compared to the nominal channel our safety channel relies on simpler algorithms and fewer safety sensors to maneuver the vehicle, which reduces the development costs for introducing redundancy. Furthermore, this pattern does not assume additional hardware for functional redundancy, which helps reducing production costs. Also, by distributing the safety responsibility to the VSM layer, the function controllers require lower integrity, which contributes to reducing associates costs.

# 4 FAULT DETECTION AND DIAGNOSIS

Thanks to its distributed nature, the DSM can effectively detect and diagnose various fault models by monitoring SoCs, networks and distributed software of the automated driving system. In this section, we discuss responsibilities of the DSM's layers in monitoring of fault models. Table I presents examples of fault models accompanied with monitoring techniques that the DSM layers can use to detect system malfunctioning. The system components column covers the whole vehicle infrastructure for automated driving, including sensors, actuators, processors, and networks. Examples of fault models and performance limitations are classified according to the DFA (Dependent Failure Analysis) coupling factor classes from [1]. Performance limitations from [2] provoke hazards when the fault-free AD system cannot safely handle a road situation, for example, because of machine learning implementation restrictions. Note that the presented fault model is not comprehensive, but rather exemplify areas of responsibility of the DSM layers. Furthermore, the monitoring techniques column includes examples of how the system can detect or mitigate corresponding fault models. Several monitoring techniques, such as heartbeats and message rate monitoring, depend on the middleware software for distributed coordination of the DSM and AD functions. Finally, the responsible DSM layer is specified along with an indication of the minimum required safety integrity level.

TABLE I. ALLOCATION OF EXAMPLE FAULT MODELS AND MONITORING TECHNIQUES TO THE DSM LAYERS

| system component | example fault model and performance limitations | failure type | example monitoring or mitigation technique | DSM layer responsibility | DSM layer integrity |
|---|---|---|---|---|---|
| sensors | message rate jitter | communication | message rate monitoring | SFM, safety of AD functions | low |
| | wrong ego-vehicle localization | component failure | fusion with heterogeneous sensors, comparison with safety channel | | |
| | missed or ghost object, or incorrect location of a detected object | component failure | fusion with heterogeneous sensors, comparison with safety channel, ODD checker | | |
| AD functions (localization, perception, prediction, planning, control) | incorrect world model perception | component failure | comparison with safety channel | | |
| | incorrect ego-vehicle localization after sensor fusion | component failure | comparison with safety channel | | |
| | path planning incurs collision | component failure | comparison with safety channel | | |
| | incorrect prediction for non-ego vehicle | component failure | comparison with safety channel | | |
| | vehicle actuator command drives vehicle off-road | component failure | comparison with safety channel | | |
| | vehicle actuator command results in loss of vehicle control | component failure | high-integrity (software) design | | |
| | application deadlock or crash | component failure | heartbeats, watchdogs | | |
| function controllers (SoCs) | memory corruption | shared resource | error-correction and detection codes, memory BIST | CSM, safety of the function controller hardware and platform software (hypervisor, OS, firmware) | medium |
| | processor core failure | shared resource | software self-test, logic BIST | | |
| | OS kernel panic, driver crash, hard fault | shared resource | heartbeats, watchdogs | | |
| | firmware crash or deadlock | shared resource | diagnostic messages with watchdog | | |
| | out-of-memory error or memory leak | shared resource | diagnostic messages with watchdog | | |
| hypervisor | CPU scheduler is stuck | shared resource | heartbeats, watchdogs | | |
| | wrong VM scheduling | shared resource | diagnostic messages with watchdog | | |
| | isolation failure (e.g. TLB failure) | component failure | software self-test, logic BIST | | |
| DSM SFM | arbitrary fault (timing, message content) | component failure | diagnostic messages with watchdog | | |
| DSM VSM | silence or wrong diagnostic response | component failure | diagnostic messages with watchdog | | |
| safety controller (SoC) | memory corruption | shared resource | error-correction and detection codes, memory BIST | VSM, vehicle safety, including monitoring of data and power networks integrity | high |
| | processor core failure | shared resource | software self-tests, logic BIST | | |
| | on-chip IP failure | component failure | on-chip safety monitors | | |
| | on-chip network failure | component failure | on-chip safety monitors | | |
| | SoC IO interface failure | component failure | on-chip safety monitors | | |
| | software fault models | component failure | control flow checks | | |
| actuators | no status reading | component failure | heartbeats, watchdogs | | |
| | wrong status reading (e.g. wheel speed) | component failure | high-integrity sensor design | | |
| data networks | corrupted packet | communication | heartbeats, watchdogs | | |
| | short-circuit, open wire, stuck-at fault | communication | heartbeats, watchdogs | | |
| | message timing violation | communication | packet transmission monitoring | | |
| | message priority violation | communication | packet transmission monitoring | | |
| power network | outage, undervoltage, overvoltage, short-circuit | systematic coupling | power management IC status checks | | |
| DSM CSM | arbitrary fault (timing, message content) | component failure | challenge-response with watchdog | | |

The primary responsibility of the SFM layer is to monitor the AD and ODD sensors and AD functions, such as localization, perception, and path planning. Modern smart sensors integrate complex processing and networking onboard. Therefore, their failure modes span from communication issues (e.g. message publish rate jitter), to software issues (e.g. wrong ego-vehicle localization), to performance limitations of the algorithms (e.g. ghost and missed objects [29]). ECUs, SoCs and VMs running AD functions will exhibit similar fault models. Examples of monitoring and mitigations techniques include message rate monitoring, data fusion of multi-modal sensors, and comparison against an independent safety channel. Furthermore, the SFM layer can perform fast sophisticated diagnosis of the health state of the AD functionality thanks to its proximity to the FUN layers running in the same powerful VM. For example, it can track timing and priority of the published packets, examine packet content using in-range and out-of-range checks, inspect drivers, the OS kernel, etc.

The CSM layer is responsible for monitoring faults in the function controllers (SoCs) and VSM. Each SoC and the low-level software (e.g. hypervisor) can suffer from various failure modes: on-chip memory corruption, stuck-at faults in the processor core and NoCs, hypervisor isolation failure, firmware deadlock or priority inversion. These fault models often occur in shared SoC resources and affect multiple components at the higher level of the AD system functionality. Therefore, handling of these fault models requires a higher integrity level than that of the SFM layer. Besides mentioned in Table I monitoring techniques the CSM can deploy on-chip hardware monitors for clock jitter, IO errors, control flow monitoring and lock-step processing and data transfer.

In our DSM the VC CSM monitors all the other NVC function controllers using a challenge-response protocol [20]. The challenge-response protocol covers memory- and processor-related fault models on top of traditional liveliness heartbeats with a watchdog. Furthermore, the CSM also serves as a fall-over mechanism when the VSM layer on the safety controller fails. To detect a fail-silent failure of the VSM, the CSM can simply monitor the VSM heartbeats using a watchdog. The CSM can safely maneuver the vehicle only by cooperating with all the fully operational FUN layers implementing the AD functionality.

Without the fully operational FUN layers, the CSM can only send basic emergency stop [12] commands to the vehicle actuators via an emergency channel. Besides virtualization and isolation, the implementation of hypervisors on the function controllers allows the CSM layer to quickly pause the VMs if a fault or a hazardous situation is detected.

The overall vehicle safety is managed by the high integrity VSM layer. Besides monitoring its own safety controller similar to the CSM, it is responsible for checking the health of the VC CSM, actuators, data and power networks. Steering, braking, and acceleration actuators share their (health) status similar to smart sensors. This status is continuously analyzed by the VSM. Furthermore, the VSM actively monitors the data networks by observing packet transmissions and querying network gateways and switches. Besides checking for unmet real-time deadlines, the VSM can analyze message priorities and content. Finally, the VSM can control the power distribution network and power management ICs to power off the VC CSM, sensors, and networks, when necessary.

It is noteworthy that besides checking AD system functionality, the DSM monitors its own components for malfunctioning and has several fail-over degraded modes to cope with faults in all its layers. The multiple SFM and CSM layers are allowed to fail arbitrarily and, hence, have only low and medium safety integrity requirements, which helps reduce the overall development cost, silicon area and power consumption. However, the VSM layer with modest computation and networking requirements does need a high integrity level to reliably manage other safety layers in the AD system.

## 5 DEGRADED MODES OF OPERATION

When the AD system is in the fault-free Nominal mode, the VC FUN layer is controlling the vehicle. All the safety layers are performing different monitoring tasks at the same time. Once a fault is detected by any of the safety layers, the DSM activates a degraded operation mode based on the current AD system status.

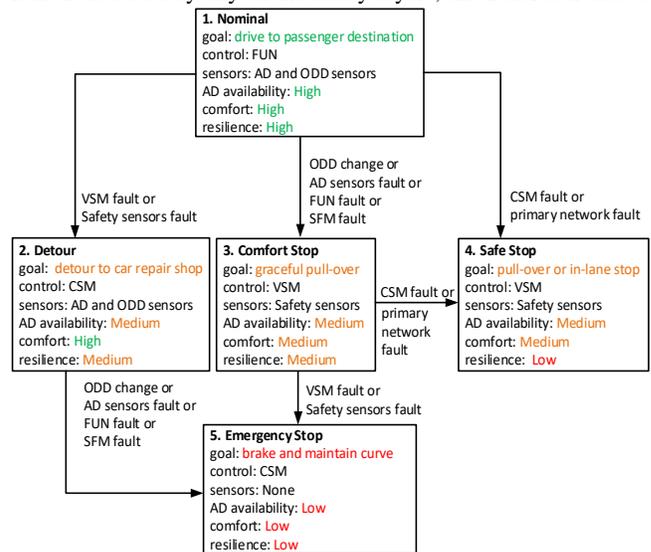

**Figure 2: DSM degradation policy**

We consider five operating modes in our DSM: `Nominal`, `Detour`, `Comfort Stop`, `Safe Stop`, and `Emergency Stop`. The degradation policies between the modes are shown in Figure 2. In this figure the labels on each transition represent the trigger of the degradation; "AD availability" represents the availability of the AD system functionalities; "comfort" indicates the passenger comfort which depends on the AD availability during the operation; and "resilience" denotes the fault tolerance of the vehicle. In the following subsections we describe the DSM behavior in each operation mode.

## 5.1 Nominal Mode

Figure 3 illustrates the behavior of the AD system with our DSM in the fault-free `Nominal` mode. In this mode all the AD system components are fully functional. The vehicle is controlled by the `VC FUN` layer in the nominal channel. Meanwhile the safety channel is on hot standby. The `VSM` layer receives and monitors the safety sensor data via the secondary network channel but does not send any control command to the actuators. All types of monitoring between safety layers are going on as well:

a) All the AD sensors, ODD sensors, and `FUN` layers are monitored by the corresponding `SFM` layers. The safety sensors are monitored by the `VSM` layer.

b) Each `SFM` layer and the NoC is monitored by the `CSM` layer on the same function controller.

c) All `NVC` function controllers are monitored by the `VC CSM` layer (not shown in Figure 3 for simplicity).

d) The `VSM` layer monitors both networks.

e) The `VC CSM` and the `VSM` layers monitor each other using the challenge-response protocol [20] through the primary network.

## 5.2 Detour Mode

The AD system nominal channel is fully functional in the `Detour` mode. Upon detecting a fault in the safety channel, the `VC CSM` layer is able to make the vehicle detour to the nearest car repair shop without sacrificing passenger comfort. Since the monitoring activities in the nominal channel works properly, the AD system can further degrade to the `Emergency Stop` mode if faults or hazardous situation are detected in the nominal channel as illustrated in Figure 2.

## 5.3 Comfort Stop Mode

The AD system safety channel is fully functional but only safety sensors are used in the `Comfort Stop` mode. The `VSM` layer in the safety channel can control the vehicle to do a graceful pull-over and guarantee sufficient passenger comfort. As Figure 2 shows, while in this mode the AD system can degrade further to the `Safe Stop` mode when the `VSM` detects a fault in the primary network or the `VC CSM` or to the `Emergency Stop` mode when `VC CSM` detects a `VSM` fault.

## 5.4 Safe Stop Mode

In this mode the AD system safety channel is still fully functional, but the nominal channel cannot perform any degradation mode anymore due to the `VC CSM` fault or primary network fault as shown in Figure 2. Because we assume components on the function controller are fail-arbitrary, to perform the safe stop the `VSM` first powers off the `VC` function controller to prevent it producing any arbitrary output, then it starts the `Safe Stop` procedure to quickly pull over the vehicle or even performs an in-lane stop when necessary. We assume in this mode the vehicle is immediately in a safe state.

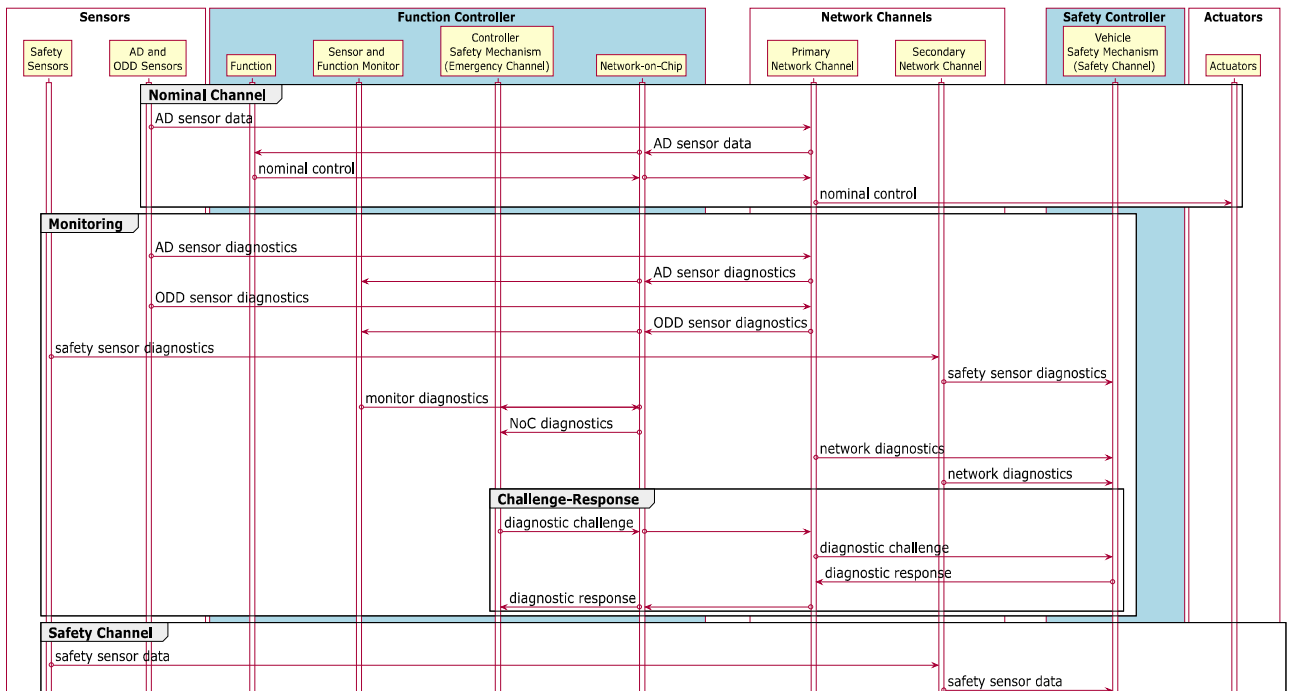

**Figure 3: Fault-free nominal mode of the Distributed Safety Mechanism**

## 5.5 Emergency Stop Mode

The `Emergency Stop` mode can be reached by degrading from the `Detour` mode or the `Comfort Stop` mode. In this mode, the AD system nominal channel is only partially functional, which means the VC `CSM` does not have all information needed for the advanced `Detour` operation but can still control the vehicle via the primary network. In addition, the AD system safety channel in the `Emergency Stop` mode cannot perform any degradation anymore due to the `VSM` or safety sensor faults as shown in Figure 2. Therefore, the VC `CSM` simply sends braking commands to the vehicle actuators to blindly stop the vehicle as quickly as possible. Obviously, the passenger comfort cannot be guaranteed during this procedure. We assume the vehicle is immediately in a safe

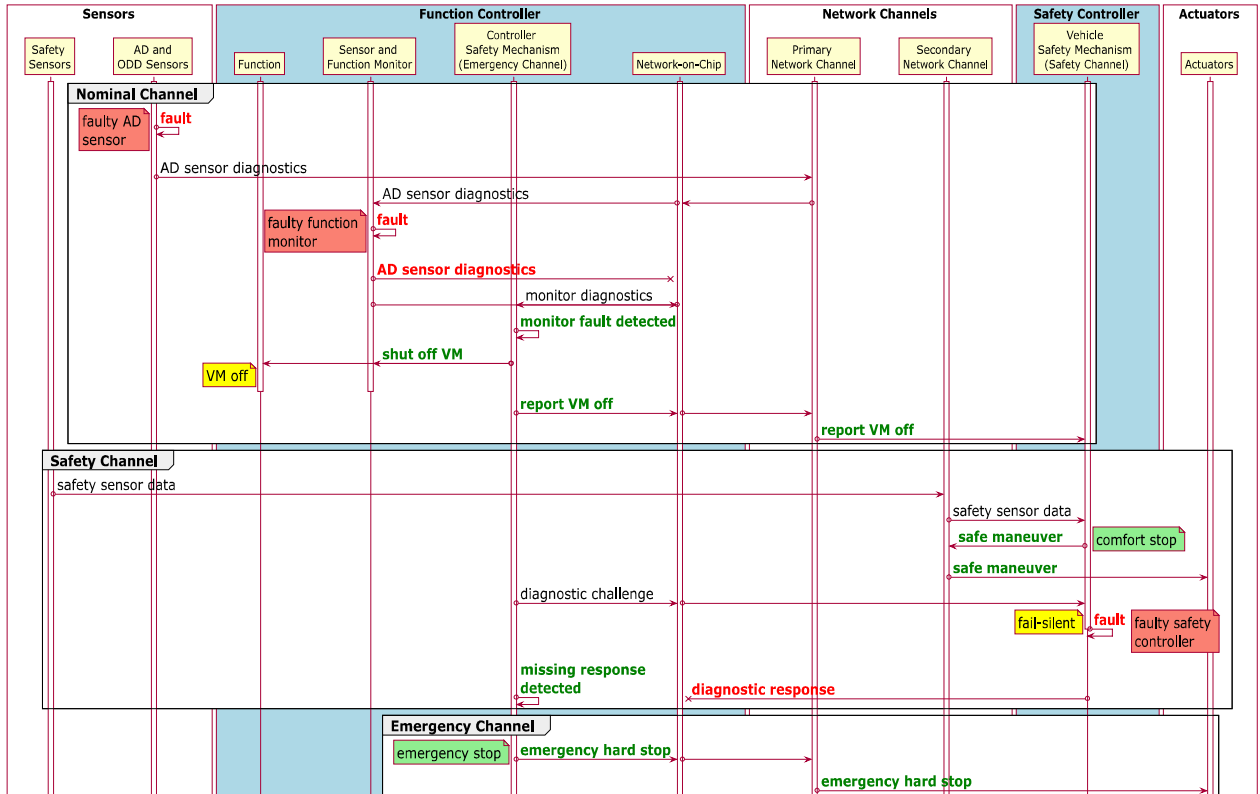

Figure 4: Triple-point fault handling: first an AD sensor fault then a function monitor fault then a vehicle safety mechanism fault

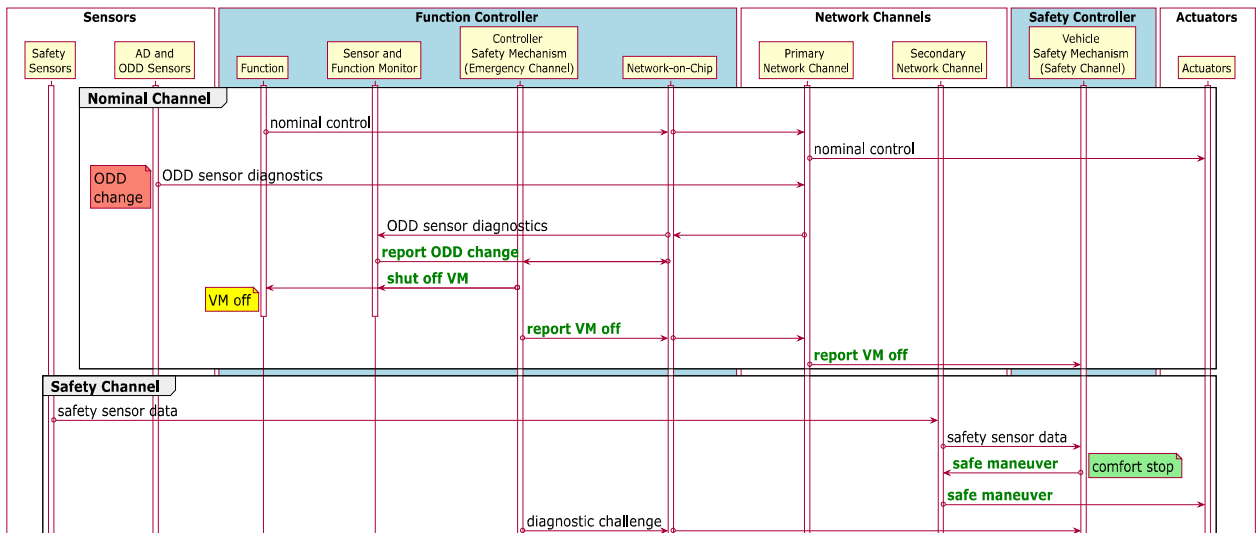

Figure 5: Handling of a SOTIF scenario when leaving the operational design domain

state in this mode.

## 5.6 Example Use Cases

As shown in Figure 2, the AD system can be degraded from the Nominal mode first to the Comfort Stop mode, then to the Emergency Stop mode. Such a use case is shown in Figure 4. At first a fault occurs in an AD sensor. Note that the fail-arbitrary AD sensor can provide arbitrary output to the FUN layer which results in an arbitrary FUN control output. But the faulty control can reach the vehicle actuators without violating the safety goal if it is quickly suppressed by the safety mechanism. In this case, the AD sensor fault is detected by the SFM layer, but then a second fault occurs in the SFM layer before it is able to report to the CSM layer. Nevertheless, the CSM layer detects the SFM layer malfunction and reacts to it by shutting down the vehicle control VM where the faulty SFM layer is running, then reports the VM state to the VSM layer. Then the VSM layer degrades the AD system to the Comfort Stop mode. Before the vehicle reaches the safe state, however, a third fault occurs in the VSM layer. Thanks to the challenge-response monitoring technique, the CSM layer detects the VSM layer fail-silent fault and further degrades the AD system to the Emergency Stop mode.

We demonstrate how the DSM is capable to handle a SOTIF [2] scenario in Figure 5. In this case, the AD system degrades from the Nominal mode to the Comfort Stop mode when there is no systematic fault, but the ODD checker on the SFM layer detects that the ODD has changed and adequate performance of the AD system cannot be guaranteed.

## 6 MODEL AND FORMAL VERIFICATION

With growing complexity of the AD system, traditional safety analysis such as Failure Mode and Effects Analysis (FMEA), FTA, as well as road testing and simulation methods become insufficient to guarantee the correctness of the AD system, because it is infeasible to exhaustively analyze or test the numerous system states. Modeling and formal verification, however, can exhaustively and mathematically prove the desired safety properties of the AD system model. This helps reduce specification and requirement errors in the early design phase. Furthermore, certifiable code can be generated from the formally verified model. The executable model can serve as a golden reference throughout the product life-cycle and help clarify communication among design stakeholders.

In this work we modeled the DSM architecture shown in Figure 1 using the mCRL2 formal specification language [24] and formalized the safety requirements of the model in the modal μ-calculus [25]. Each and every component in Figure 1 is modeled as a Finite State Machines (FSM). The FSM transitions specify the DSM behavior in monitoring and fault handling procedures, which is illustrated in Figure 2 with the high-level degradation policy diagram and in sequence diagrams in Figure 3 to Figure 5.

The model consists of 746 lines of mCRL2 code and 71 lines of μ-calculus formulas. The mCRL2 model is also executable and its behavior can easily be simulated in the mCRL2 simulator.

Table II shows the state space of each FSM in our DSM model. The AD system model has all these FSMs run in parallel, resulting in a state space of more than 2 million states and 2 billion transitions. The mCRL2 tool is able to mathematically prove the properties of this model while it is intractable to exhaustively test all these transitions using traditional road testing or AD simulation setup. In the following subsections we present details of the DSM model.

TABLE II. STATE SPACE OF DSM STATE MACHINE

| State Machine | # states | # transition triggers | # transitions |
|---|---|---|---|
| AD/ODD sensors | 3 | 4 | 3 |
| Safety sensors | 3 | 4 | 3 |
| FUN | 6 | 9 | 32 |
| SFM | 8 | 10 | 58 |
| CSM | 10 | 23 | 167 |
| VSM | 10 | 10 | 39 |
| Actuators | 5 | 6 | 23 |

## 6.1 Model Specifications

The AD system components communicate using the software middleware stacks implementing the publish-subscribe pattern [19]. Thus, it is essential to properly model the publish-subscribe communication behavior. In our DSM model, information is shared as messages. Any participant on the same network can receive messages by subscribing to a message topic. We model two types of networks in the DSM, namely the NoC and the off-chip networks. The DSM layers communicate using three message types:
1) Sensor Data Messages. These messages contain the sensor data communicated in the AD system.
2) Vehicle Control Command Messages. These control commands are sent to the vehicle actuators by the VC FUN layer, the VC CSM layer, or the VSM layer.
3) Diagnostic Messages. These messages report the status of the monitored component.

All the messages consist of an action carrying four identification parameters. For example, the action of the vehicle control CMS layer sending a diagnostic message via the primary network channel reporting the vehicle control VM being shut off is modeled as: send(VC, VM_VC, VC_VM_OFF, primary).

The four parameters from left to right are explained in Table III. For sensor data communication, cid and vid identify the receiver component. For power management actions, network is set to power. Actions modeled using the same structure are send and receive for communication via the off-chip networks, and NoCsend and NoCreceive for communication via the NoC.

TABLE III. MESSAGE PARAMETERS AND EXAMPLES

| Parameter | Description | Parameter examples |
|---|---|---|
| cid | controller ID or the name of other component | FUN, SFM, VC (short for VC CSM), NVC (short for NVC CSM), VSM, SAFE_SENSOR, NETWORK |
| vid | virtual machine ID | VM_VC (the VM running VC FUN), NULL (if the SoC runs no VM), VM_1, VM_2, etc. |
| info | message content or environmental fault identification | VC_VM_OFF, NOMINAL, Sens_OK, EMERGENCY_STOP, FUN_FAULT, |
| network | network channel identification | primary, secondary, NoC, power |

All the AD system behavior is modeled by actions in the model. We present a few examples here:
- network_ingress, network_egress: The ingress and egress phases of the network communication.
- fault_injection_*: These actions model the occurrences of faults. The * can be replaced by any of the target components. The events in Figure 4 and Figure 5 next to the red labels are of this type.
- vc_csm_shuts_off_vc_vm: These actions model the VC CSM shutting off the VC VM it manages. It is illustrated in Figure 4 as "shut off VM".
- vsm_powers_off_vc_controller: This action models the VSM layer powering off the VC CSM SoC.

### 6.2 Model Assumptions

We made the following model assumptions to clarify the scope of our study and keep the model small enough [25] to be verifiable while maintaining model's relevance:

a) All faults are permanent, atomic, not safe, and always detected. Fault detection can be improved with adequate monitoring techniques without compromising the model. Transient faults are out of scope in this work.

b) NoCs have infinite capacity, transfer data atomically and never fail. Note, faults in NoCs can be detected by different monitors in the AD system thanks to the layered architecture in our DSM concept. But fault handling in NoCs is part of our future work.

c) The power supply never fails. Power loss in parts of the AD system can be detected by various methods like heartbeat monitoring or challenge-response mechanism and therefore can be mitigated with degraded modes thanks to the layered structure of the DSM concept. However, power management is in general out of our scope.

d) The secondary network never fails and is only accessible by the safety sensors, VSM, and the actuators. We assume the secondary network in the safety channel is more reliable than the nominal channel due to its different and simpler implementation with limited functionalities.

e) All DSM transitions are instantaneous and do not fail.

f) VSM, safety sensors, and both off-chip networks are fail-silent, all the other components are fail-arbitrary.

g) Communications through the off-chip primary and secondary networks are split into ingress and egress phases and have a buffer of 3 messages. The egress messages are sent out in random order. We set the buffer size to 3 to model the out-of-order ingress and egress data flows and keep the state space small.

h) Only certain combinations of multiple-point faults are allowed as illustrated in Figure 2, such that there is always a degraded mode available to bring the vehicle to a safe state.

### 6.3 Safety Requirements and Formal Properties

The DSM model safety requirements are listed below:
Req 1 The AD system is deadlock-free;
Req 2 There is always one and only one non-faulty component in control of the vehicle;
Req 3 The AD system Nominal mode always degrades to Detour, Comfort Stop or Safe Stop when necessary;
Req 4 Detour must degrade to Emergency Stop when necessary;
Req 5 Comfort Stop must degrade to Safe Stop or Emergency Stop accordingly when necessary.

The DSM safety requirements are formalized using modal μ-calculus formulas [24]. Below we show two examples:
Req 1 [**true***]<**true**>**true**
Req 4 [**true***.network_egress(VC, VM_VC, DETOUR, primary)]
    [**true***] **forall** cid: CTRLR_ID, vid: VM_ID.
    [(fault_injection_fun(cid, vid, FUN_FAULT, NoC) ||
     fault_injection_sfm(cid, vid, SFM_FAULT, NoC) ||
     fault_injection_csm(cid, NULL, CSM_FAULT, NoC) ||
     fault_injection_sensor(cid, vid, SENS_FAULT, NoC))]
    [!network_egress(VC, NULL, EMERGENCY_STOP, primary)*]
    <**true***.network_egress(VC, NULL, EMERGENCY_STOP,primary)
    >**true**

### 6.4 Verification Results

The modeling and verification processes help to reduce specification errors in the DSM design which are hard to detect manually or in road testing. For example, an early version of our DSM model with network buffer size set to 3 has a state space of 215 million states and 2.92 billion transitions. We found a deadlock in this model after running the verification process using the mCRL2 toolset [24] for more than 10 consecutive days. Noteworthy, when the network buffer size was set to 1 and 2, the same model was verified to meet all the safety requirements. We debugged the issue by stepping through the FSMs transitions in different fault handling procedures. With the help of the mCRL2 simulator, we quickly discovered that the deadlock occurred when the primary network was fully occupied by the diagnostics about the safety sensor fault sent by the VSM, meanwhile the VC CSM detected the VSM fault and tried to activate the Detour mode via the fully occupied primary network. In the previously verified models, however, the diagnostics were not buffered long enough to trigger the deadlock due to the smaller size of the buffer. The issue was resolved by adding missing transitions in the VC CSM process to allow it to receive diagnostics from the VSM layer when the vehicle control command cannot be sent out yet.

The final DSM model configuration had two function controllers each running two VMs and the network buffer size set to three, resulting in a state space of over 1 billion states and 10 billion transitions. All the safety requirements were verified for the final model, which took several days of runtime on a powerful server. Although there is always a reality gap between the formal model and the actual system implementation, the formal approach helps reduce specification errors by mathematically and exhaustively proving the correctness of the arbitration logic in the early design phase prior to the costly AD system implementation and in-field testing.

# 7 CONCLUSION

Our work presents a fail-operational safety concept combining safety measures for fault handling, performance limitations mitigation and driving decision making. The adopted safety mechanisms handle multiple-point faults in both the AD system and in the safety mechanisms themselves. The proposed AD system architecture is made cost-effective by using mostly fail-arbitrary components as well as additional emergency and safety channels which are simpler than the nominal channel. Based on the AD system health status and situational awareness, the DSM running on multiple SoCs and virtual machines activates various degraded modes of automated driving to keep the vehicle as safe as possible. Furthermore, we present an example allocation of fault models to our safety mechanism layers and discuss monitoring techniques to detect faults and ODD changes. To reduce specification errors, we modeled our DSM arbitration logic in the mCRL2 language and formally verified five safety requirements in μ-calculus. Our formally verified highly resilient safety concept can serve as a holistic design pattern for AD system minimizing road fatalities and reducing costly testing.

Promising future research directions include modeling timing properties, computation of failure rates using probabilistic models, handling of transient faults and quantification of cost-effectiveness of the presented safety concept.